\documentclass[10pt,twocolumn,letterpaper]{article}

\usepackage{cvpr}
\usepackage{times}
\usepackage{epsfig}
\usepackage{graphicx}
\usepackage{amsmath}
\usepackage{amssymb}
\usepackage{float}
\usepackage{mathtools}
\usepackage{subfig}
\usepackage{graphicx}
\usepackage{booktabs}
\usepackage{dsfont}
\usepackage{enumitem}
\usepackage{caption}

\DeclarePairedDelimiter\abs{\lvert}{\rvert}%
\DeclarePairedDelimiter\norm{\lVert}{\rVert}%

\usepackage[pagebackref=true,breaklinks=true,letterpaper=true,colorlinks,bookmarks=false]{hyperref}

\usepackage[usenames, dvipsnames]{color} 
\definecolor{purple}{rgb}{0.5, 0.0, 0.5}
\definecolor{orange}{rgb}{1, 0.65, 0}
\definecolor{lightgreen}{rgb}{0.68, 1, 0.18}
\definecolor{darkgreen}{rgb}{0.09, 0.32, 0.24}
\definecolor{darkred}{rgb}{0.6, 0, 0}
\definecolor{brown}{rgb}{0.64, 0.16, 0.16}
\iffalse 
  \newcommand{\tung}[1]{\noindent}
  \newcommand{\corina}[1]{\noindent}
  \newcommand{\freddy}[1]{\noindent}
  \newcommand{\eric}[1]{\noindent}
  \newcommand{\oscar}[1]{\noindent}
  \newcommand{\todo}[1]{\noindent}
\else
  \newcommand{\tung}[1]{\textcolor{blue}{\bf [TP: #1]}}
  \newcommand{\corina}[1]{\textcolor{orange}{\bf [ECG: #1]}}
  \newcommand{\freddy}[1]{\textcolor{purple}{\bf [FB: #1]}}
  \newcommand{\eric}[1]{\textcolor{brown}{\bf [EW: #1]}}
  \newcommand{\oscar}[1]{\textcolor{darkgreen}{\bf [OB: #1]}}
  \newcommand{\todo}[1]{\textcolor{red}{\bf [Todo: #1]}}
\fi

\newcommand{\hitrate}[0]{HitRate\textsubscript{5, 2m}\xspace}
\newcommand{\ade}[1]{minADE\textsubscript{#1}\xspace}

\cvprfinalcopy 

\def\cvprPaperID{8765} 

\begin{document}

\title{CoverNet: Multimodal Behavior Prediction using Trajectory Sets}

\author{Tung Phan-Minh\thanks{Work done during an internship at nuTonomy, an Aptiv company.}\\
Caltech\\
{\tt\small tung@caltech.edu}
\and
Elena Corina Grigore, Freddy A. Boulton, Oscar Beijbom, and Eric M. Wolff\\
nuTonomy, an Aptiv company\\
{\tt\small \{elena.corina.grigore, freddy.boulton, oscar, eric\}@nutonomy.com}
}

\maketitle

\begin{abstract}
We present CoverNet, a new method for multimodal, probabilistic trajectory prediction for urban driving. Previous work has employed a variety of methods, including multimodal regression, occupancy maps, and 1-step stochastic policies. We instead frame the trajectory prediction problem as classification over a diverse set of trajectories. The size of this set remains manageable due to the limited number of distinct actions that can be taken over a reasonable prediction horizon. We structure the trajectory set to a) ensure a desired level of coverage of the state space, and b) eliminate physically impossible trajectories. By dynamically generating trajectory sets based on the agent's current state, we can further improve our method's efficiency. We demonstrate our approach on public, real-world self-driving datasets, and show that it outperforms state-of-the-art methods.
\end{abstract}


\begin{figure*}[htp]
\begin{center}
   \includegraphics[width=1.0\linewidth]{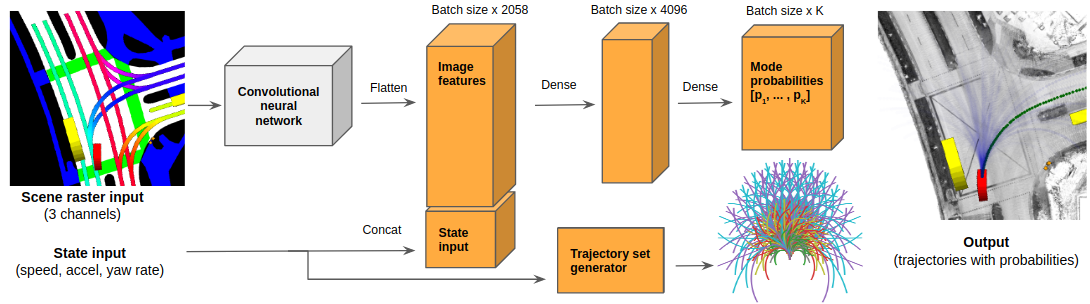}
\end{center}
   \caption{CoverNet overview. We generate a trajectory set (fixed or dynamic based on current state) that we classify over. The input and backbone follow~\cite{cui2019multimodal}.}
\label{fig:network-architecture}
\end{figure*}

\section{Introduction}
We are motivated by autonomous systems operating in dynamic, interactive, and uncertain environments. Specifically, we focus on the problem of a self-driving car navigating in an urban environment, where it must share the road with a diverse set of other agents, including vehicles, bicyclists, and pedestrians. In this context, reasoning about the possible future states of agents is critical for safe and confident operation. Effective prediction of future agent states depends on both road context (e.g., lane geometry, crosswalks, traffic lights) and the recent behavior of other agents.

Trajectory prediction is inherently challenging due to a wide distribution of agent preferences (e.g., a cautious vs. aggressive) and intents (e.g., turn right vs. go straight). Useful predictions must represent multiple possibilities and their associated likelihoods. Furthermore, we expect that predicted trajectories are physically realizable.

Multimodal regression models appear naturally suited for this task, but may degenerate during training into a single mode. Avoiding this ``mode collapse'' requires careful considerations~\cite{cui2019multimodal,chai2019multipath,hong2019zoox}. Additionaly, most state-of-the-art methods predict unconstrained positions~\cite{cui2019multimodal,chai2019multipath,hong2019zoox,rhinehart2019precog}, resulting in trajectories that may not be physically possible for execution (\cite{cui2019deep-kinematic} is a recent exception). Our main insights leverage domain-specific knowledge to effectively structure the output representation and address these concerns.

Our first insight is that there are relatively few \emph{distinct} actions that can be taken over a \emph{reasonable} time horizon. Dynamic constraints considerably limit the set of reachable states over a standard six second prediction horizon, and the inherent uncertainty in agent behavior outweighs small approximation errors. We exploit this insight to formulate multimodal, probabilistic trajectory prediction as classification over a trajectory set. This avoids mode collapse and lets the user design the trajectory set to meet specific requirements (e.g., dynamically feasible, coverage guarantees).

Our second insight is that predicted trajectories should be consistent with the current dynamic state. Thus, we formulate our output as motions relative to our initial state (e.g., turn slightly right, accelerate). When integrated with a dynamics model, the output is converted to an appropriate sequence of positions. Beyond helping ensure physically valid trajectories, this \emph{dynamic} output representation ensures that the outputs are diverse in the control space across a wide range of speeds. While~\cite{cui2019deep-kinematic} exploit a similar insight for regression, we extend the use of a dynamic representation to classification and anchor-box regression.

We now summarize our main contributions on multimodal, probabilistic trajectory prediction with CoverNet:
\begin{itemize}[nosep]
    \item introduce the notion of trajectory sets for multimodal trajectory prediction, and show how to generate them in both a fixed and dynamic manner;
    \item compare state-of-the-art methods on nuScenes~\cite{nuscenes2019}, a public, real-world urban driving benchmark;
    \item empirically show the benefits of classification on trajectory sets over multimodal regression.
\end{itemize}

\section{Related Work}
\label{sec:related-work}
We focus on trajectory prediction approaches based on deep learning, and refer the reader to ~\cite{lefevre2014survey} for a survey of more classical approaches. The approaches below typically use CNNs to combine agent history with scene context, and vary significantly in their output representations. Depending on the method, the scene context will include everything from the past states of a single agent, to the past states of all agents along with high-fidelity map information.

Stochastic approaches encode choice over multiple possibilities via sampling random variables. One of the earliest works on motion forecasting frames the problem as learning stochastic 1-step policies~\cite{kitani2012activity-forecasting}. R2P2~\cite{rhinehart2018r2p2} improves sample coverage for such policies via a symmetric KL loss. Recent work has considered the multiagent setting~\cite{rhinehart2019precog} and uncertainty in the model itself~\cite{henaff2019mpc}. Other methods generate samples using CVAEs~\cite{hong2019zoox, lee2017desire, bhattacharyya2018best-of-many, ivanovic2018multimodal-human} or GANs~\cite{sadeghian2019sophie-gan, gupta2018social-gan, zhao2019MultiAgentTF}. Stochastic approaches can be computationally expensive due to a) repeated 1-step rollouts (in the 1-step policy approach), or b) requiring a large number of samples for acceptable performance (often hard to determine in practice).

Unimodal approaches output a single trajectory per agent ~\cite{luo2018fast-and-furious, casas18intent-net, djuric2018shortterm, alahi2016social-lstm}. This is often unable to adequately capture possibilities in complex scenarios, even when predicting Gaussian uncertainty. These methods typically average over behaviors, which may result in nonsensical trajectories (e.g., halfway between a right turn and going straight). 

Multimodal approaches output either a distribution over multiple trajectories ~\cite{chai2019multipath, cui2019multimodal, hong2019zoox, deo2018conv-pool-vehicles} or a spatial-temporal occupancy map ~\cite{hong2019zoox, bansal2019chauffeurnet, zeng2019costmap}. The latter flexibly captures multiple outcomes, but often has large memory requirements for grids at reasonable resolutions. Sampling trajectories from an occupancy map a) is not well defined, and b) adds additional compute during inference. Multimodal regression methods can easily suffer from ``mode collapse'' to a single mode, leading~\cite{chai2019multipath} to use a fixed set of anchor boxes. In contrast, the strength of our contribution lies in framing the problem as classification rather than regression. We also contribute three methods of creating trajectory sets to classify over, and achieve performance improvements over~\cite{chai2019multipath}.

Most trajectory prediction methods do not explicitly encode motion constraints, predicting trajectories that can be physically infeasible (a recent exception is~\cite{cui2019deep-kinematic}). By careful choice of our output representation, we exclude all trajectories that would be physically impossible to execute. Although our predictions can result in off-road trajectories at test time, our model learns to assign them a low probability as long as such trajectories are not included during training.

Graph search is a classic approach to motion planning~\cite{lavalle2006planning-algos}, and often used in urban driving applications~\cite{buehler2009darpa-urban-challenge}. A motion planner grows a compact graph (or tree) of possible motions, and computes the best trajectory from this set (e.g., max clearance from obstacles). Since we do not know the other agent's goals or preferences, we cannot directly plan over the trajectory set. Instead, we implicitly estimate these features and directly classify over the set of possible trajectories. There is a fundamental tension between the size of the trajectory set, and the coverage of all potential motions~\cite{branicky2008path-diversity}. Since we are only trying to predict the motions of other vehicles well enough to drive, we can easily accept small errors over moderate time horizons (3 to 6 seconds).

Comparing results on trajectory prediction for self-driving cars in urban environments is challenging. Numerous papers are evaluated purely on internal datasets~\cite{cui2019multimodal,zeng2019costmap,casas18intent-net,bansal2019chauffeurnet}, as common public datasets are either relatively small~\cite{geiger2012kitti}, focused on highway driving~\cite{colyar2007ngsim}, or are tangentially related to driving~\cite{robicquet2016stanford-drone}. While there are encouraging new developments in public datasets ~\cite{chang2019argoverse, hong2019zoox}, there is no standard. To help provide clear and open results, we evaluate our models on nuScenes~\cite{nuscenes2019}, a recent public self-driving car dataset focused on urban driving.


\section{Method}
\label{sec:method}
\label{sec:trajectory-sets}
In this section we outline the main contribution of the paper: a novel method for trajectory set generation, and show how it can be used for behavior prediction.

\subsection{Notation}
CoverNet computes a multimodal, probabilistic prediction of the future states of a given vehicle using i) the current and past states of all agents (e.g., vehicles, pedestrians, bicyclists), and ii) a high-definition map. 

We assume access to the state outputs of an object detection and tracking system of sufficient quality for self-driving. We denote the set of agents that a self-driving car interacts with at time $t$ by $\mathcal{I}_t$ and $s^i_{t}$ the state of agent $i \in \mathcal{I}_t$ at time $t$. Let $s^{i}_{m:n} = \left[s^i_{m}, \ldots, s^i_{n}\right]$ where $m < n$ and $i \in \mathcal{I}_t$ denote the discrete-time trajectory of agent $i$ from for times $t = m, \ldots, n$. 


Furthermore, we assume access to a high-definition map including lane geometry, crosswalks, drivable area, and other relevant information.

Let $\mathcal{C} = \{ \bigcup_i s^{i}_{t-m:t}; \textrm{map} \}$ denote the scene context over the past $m$ steps (i.e., map and partial history of all agents).

Figure~\ref{fig:network-architecture} overviews our model architecture. It largely follows~\cite{cui2019multimodal}, with the key difference in the output representation (see Section~\ref{sec:output}). We use ResNet-50~\cite{he2015resnet} given its effectiveness in this domain~\cite{cui2019multimodal,chai2019multipath}.

While our network only computes a prediction for a single agent at a time, our approach can be extended to simultaneously predict for all agents in a similar manner as~\cite{chai2019multipath}. We focus on single agent predictions (as in~\cite{cui2019multimodal}) both to simplify the paper and focus on our main contributions.

The next sections detail our input and output representations. Our innovations are in our output representations (the \emph{dynamic} encoding of trajectories), and in treating the problem as classification over a diverse set of trajectories.

\subsection{Output representation}
\label{sec:output}
Due to the relatively short trajectory prediction horizons (up to 6 seconds), and inherent uncertainty in agent behavior, we approximate all possible motions with a set of trajectories that gives sufficient coverage of the space.

Let $\mathcal{R}(s_t)$ be the set of all states that can be reached by an agent with current state $s_t$ in $N$ timesteps (purely based on physical capabilities). 
We approximate this set by a finite number of trajectories, defining a trajectory set $\mathcal{K}= \{s_{t:t+N}\}$.
We define a \emph{dynamic} trajectory set generator as a function $f_N: s_0 \to \mathcal{K}$, which allows the trajectory set to be consistent with the current dynamics. 
In contrast, a \emph{fixed} generator does not use information about the current state, and thus returns the same trajectories for each instance. 
We discuss trajectory set construction in Section~\ref{sec:trajectory-sets}.

We encode multimodal, probabilistic trajectory predictions by classifying over the appropriate trajectory set given an agent of interest and the scene context $\mathcal{C}$. As is common in the classification literature, we use the softmax distribution. Concretely, the probability of the $k$-th trajectory is given as $p(s_{t:t+N}^k | x) = \frac{\exp f_k(x)}{\sum_i\exp f_i(x)}$, where $f_i(x) \in \mathbb{R}$ is the output of the network's penultimate layer. 

In contrast to previous work~\cite{cui2019multimodal,chai2019multipath}, we choose not to learn an uncertainty distribution over the space. While it is straightforward to add Gaussian uncertainty along each trajectory in a similar manner to~\cite{chai2019multipath}, the density of our trajectory sets reduces its benefit compared to the case when there are only a handful of modes.

\begin{figure}[t]
    \centering
    \subfloat[][fixed]{%
        \includegraphics[width=0.5\linewidth]{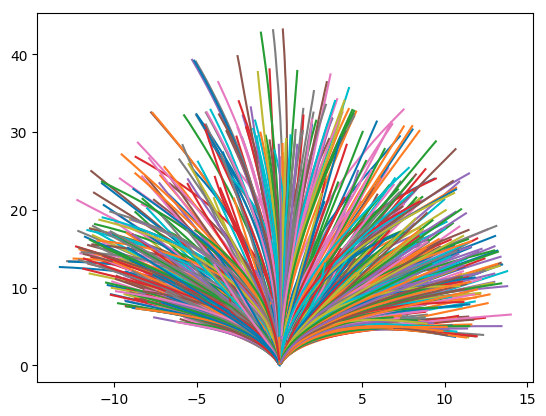}%
        }%
    \hfill%
    \subfloat[][dynamic]{%
        \includegraphics[width=0.5\linewidth]{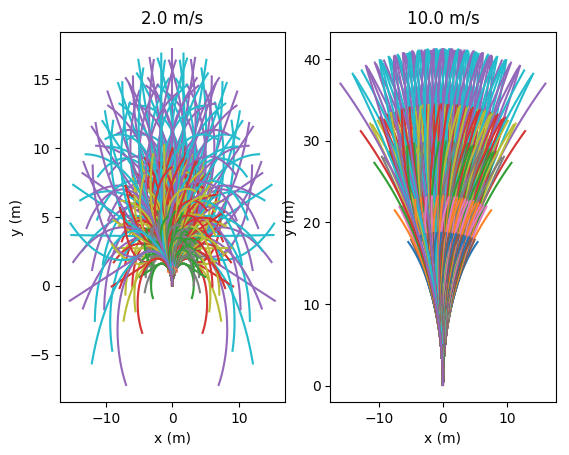}%
        \label{fig:dynamic-trajectory-set}%
        }%
    \caption{Overview of trajectory set generation approaches.}
    \label{fig:trajectory-sets}
\end{figure}

An ideal trajectory set always contains a trajectory that is close to the ground truth. We propose two broad categories of trajectory set generation functions: fixed and dynamic (see Figure~\ref{fig:trajectory-sets}). In both cases, we normalize the current state to be at the origin, with the heading oriented upwards.

\subsection{Fixed trajectory sets}
\label{subsec:trajectory-set-fixed}
We consider a trajectory set to be \emph{fixed} if the trajectories that it contains do not change as a function of the agent's current dynamic state or environment. Intuitively, this makes it easy to classify over since it allows for a fixed enumeration over the set, but may result in many trajectories that are poor matches for the current situation.

Given a set of representative trajectory data, the problem of finding the smallest fixed approximating trajectory set $\mathcal{K}$ can be cast as an instance of the NP-hard set cover problem.~\cite{cormen2009clrs-algos}.
Approximating a dense trajectory set by a sparse trajectory set that still maintains good coverage and diversity has been studied in the context of robot motion planning~\cite{branicky2008path-diversity}.
In this work, we use a coverage metric $\delta$ defined as the maximum point-wise Euclidean distance between trajectories. 
Our trajectory set construction procedure starts with subsampling a reasonably large set $\mathcal{K'}$ of trajectories (ours have size 20,000) from the training set. Selecting an acceptable error tolerance $\varepsilon$, we proceed to find the solution to: 
\begin{equation}
    \label{eq: set-cover}
\begin{aligned}
& \underset{\mathcal{K}}{\text{argmin}}
& & \abs{\mathcal{K}} \\
& \text{subject to}
& & \mathcal{K} \subseteq \mathcal{K'}, \; \\
&
&& \forall k \in \mathcal{K'}, ~ \exists l \in \mathcal{K},~ \delta(k,l) \leq \varepsilon,
\end{aligned}
\end{equation}


where $\delta(s_{t:t+N}, \hat{s}_{t:t+N}) \coloneqq \max_{\tau=t}^{t+N}{\norm{s_{\tau} - \hat{s}_{\tau}}}_2$. We refer to this metric as the maximum point-wise $\ell^{2}$ distance.

We employ a simple greedy approximation algorithm to solve $\eqref{eq: set-cover}$, which we refer to as the \emph{bagging} algorithm. 
We cherry-pick the best among candidate trajectories to place in a bag of trajectories that will be used as the covering set. 
We repeatedly consider as candidates those trajectories that have not yet been covered and choose the one that covers the most uncovered trajectories (ties are broken arbitrarily).


Standard results (without using the specialized structure of the data) show that our deterministic greedy algorithm is suboptimal by a factor of at most $\log{(\abs{\mathcal{K'}})}$ (see Chapter 35.3~\cite{cormen2009clrs-algos}). 
In our experiments, we were able to obtain decent coverage (specifically, under 2 meters in maximum point-wise $\ell^2$ distance for 6 second trajectories) with fewer than 2,000 elements in the covering set.

\subsection{Dynamic trajectory sets}
\label{subsec:trajectory-set-dynamic}
We consider a trajectory set to be \emph{dynamic} if the trajectories that it contains change as a function of the agent's current dynamic state. This construction guarantees that all trajectories in the set are dynamically feasible.

We now describe a simple approach to constructing such a dynamic trajectory set, focused on predicting vehicle motion. We use a standard vehicle dynamical model~\cite{lavalle2006planning-algos} as similar models are effective for planning at urban (non-highway) driving speeds~\cite{kong2015kinematic-mpc-self-driving}. Our approach, however, is not limited to vehicles or any specific model. The dynamical model we use is: 
\begin{align*}
\dot{x} &= v \cos \theta \\
\dot{y} &= v \sin \theta \\
\dot{\theta} &= \frac{v}{b} \tan(u_{steer}) \\
\dot{v} &= u_{accel}
\end{align*}
with states: $x$, $y$ (position), $v$ (speed), $\theta$ (yaw); controls: $u_{steer}$ (steering angle), $u_{accel}$ (longitudinal acceleration); and parameter: $b$ (wheelbase).

\begin{figure}[t]
\begin{center}
\includegraphics[width=0.9\linewidth]{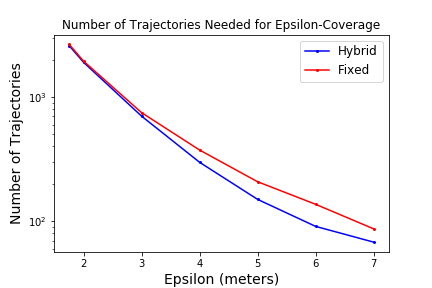}
\end{center}
\vspace{-2mm}
\caption{Number of trajectories needed for $\varepsilon$ coverage (in meters, see Section~\ref{sec:trajectory-sets})}
\vspace{-2mm}
\label{fig:epsilon-coverage}
\end{figure}

The dynamics model, controls sequence, and current state determine a trajectory $s_{t:t+N}$ by forward integration. We create a dynamic trajectory set $\mathcal{K}$ based on the current state $s_t$ by integrating forward with our dynamic model over diverse control sequences. Such a dynamic trajectory set has the possibility of being sparser than a fixed set for the same coverage, as each control sequence maps to multiple trajectories (as a function of the current state).

We parameterize the controls (output space) by a diverse set of constant lateral and longitudinal accelerations over the prediction horizon. Using lateral acceleration instead of steering angle is a way of normalizing the output over a range of speeds (a desired lateral acceleration will correspond to different steering angles as a function of speed). We convert the lateral acceleration into a steering angle assuming instantaneous circular motion \(a_{lat} = v^2 \kappa \) with curvature $\kappa = \tan({u_{steer}}) / b$. This conversion is ill-defined when the speed is near zero, so we use $\max{(v,1)}$ in place of $v$. Note that it is straightforward to expand the controls (output space) to include multiple lateral and longitudinal accelerations over a non-uniform prediction horizon.

We can further prune the dynamic trajectory set construction in a similar manner to how we handled the fixed trajectory sets in~\ref{subsec:trajectory-set-fixed}. The main difference is that the covering set here is constructed from the set of control input profiles as opposed to elements of $\mathcal{K'}$ itself. Namely, we use an analogous greedy procedure to cover the set of sample trajectories with a subset of control profiles (e.g., lateral and longitudinal accelerations as a function of time). Note that unlike the case of fixed trajectories, the synthetic nature of the dynamic profile may not guarantee 100\% coverage of $\mathcal{K'}$. To counter this problem we can also create a \emph{hybrid} trajectory set by combining a fixed and dynamic set. Particularly, we find a covering subset for the elements of $\mathcal{K'}$ that cannot be covered by the dynamic choices, and combine this subset with the dynamic choices. When the dynamic set is well-constructed, this can result in a smaller covering set as may be seen from~Figure~\ref{fig:epsilon-coverage}.


\section{Experiments}
We present empirical results on trajectory prediction of vehicles in urban environments. The following sections describe the baselines, metrics, and urban driving datasets that we considered. We used the same input representation and model architecture across our models and baselines.

\subsection{Baselines}
\noindent\textbf{Physics oracle}. We introduce a simple and interpretable model that extends classic physics-based models. We use the track's current velocity, acceleration, and yaw rate to compute the following predictions: i) constant velocity and yaw, ii) constant velocity and yaw rate, iii) constant acceleration and yaw, and iv) constant acceleration and yaw rate. The \emph{oracle} is the minimum average point-wise Euclidean distance over the four models.

\begin{figure*}
    \centering
    \subfloat[][CoverNet, fixed, $\varepsilon = 2$, 1937 modes]{%
        \includegraphics[width=0.3\textwidth]{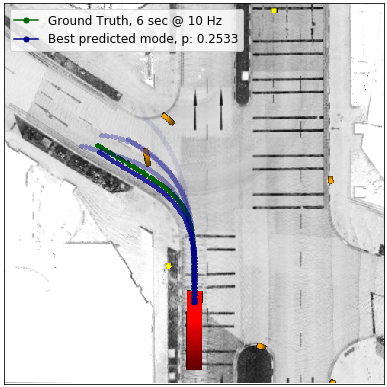}%
        }%
    \subfloat[][CoverNet, hybrid, 1024 modes]{%
        \includegraphics[width=0.3\textwidth]{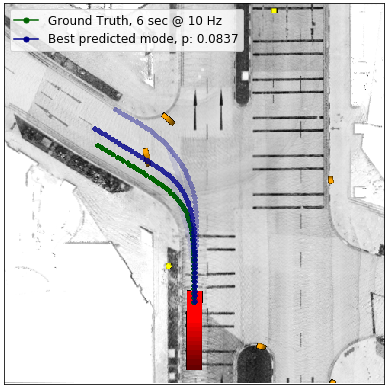}%
        \label{fig:dynamic-trajectory-set}%
        }%
    \subfloat[][CoverNet, dynamic $\varepsilon = 3$, 342 modes]{%
        \includegraphics[width=0.3\textwidth]{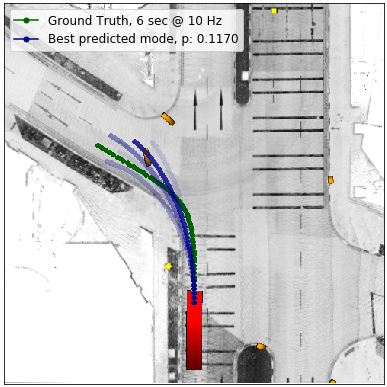}%
        \label{fig:dynamic-trajectory-set}%
        }%
    \hfill%
    \subfloat[][MTP~\cite{cui2019multimodal}, 3 modes]{%
        \includegraphics[width=0.3\textwidth]{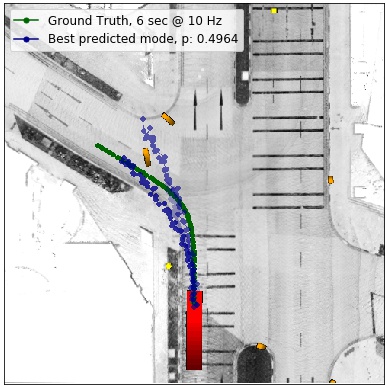}%
        }%
    \subfloat[][MultiPath~\cite{cui2019multimodal}, dynamic, 16 modes]{%
        \includegraphics[width=0.3\textwidth]{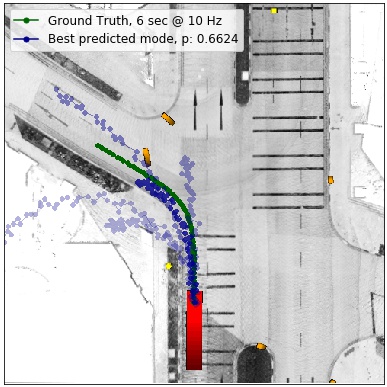}%
        \label{fig:dynamic-trajectory-set}%
        }%
    \subfloat[][MultiPath~\cite{cui2019multimodal}, 64 modes]{%
        \includegraphics[width=0.3\textwidth]{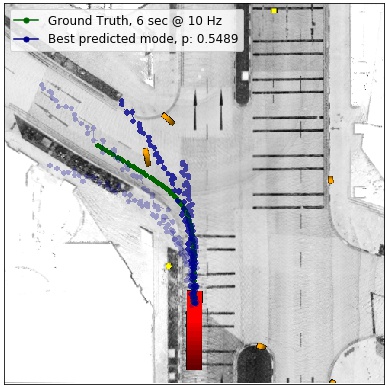}%
        \label{fig:dynamic-trajectory-set}%
        }%
    \caption{Examples of predicted trajectories on the same scene. The top row includes our CoverNet models, ranging from fixed to dynamic. The bottom row includes the baselines we compare against, as well as our dynamic templates variation. Objects in the world are rendered up to the current time.}
    \label{fig:image-examples}
\end{figure*}



\noindent\textbf{Regression baselines and extensions}.
\label{sec:regression-methods}
We compare our contribution to state-of-the-art methods by implementing two main types of regression models: multimodal regression to coordinates~\cite{cui2019multimodal} and multimodal regression to residuals from a set of anchors~\cite{chai2019multipath} (ordinal regression).
We overview these methods for completeness and to provide context for novel variations that we introduce.

\noindent\textbf{Multimodal regression to coordinates}.
Our implementation follows the details of Multiple-Trajectory Prediction (MTP)~\cite{cui2019multimodal}, adapted for our datasets. This model predicts a fixed number of trajectories (modes) and their associated probabilities. The per-agent loss (agent $i$ at time $t$) is defined as:
\begin{equation}
\label{eq:mtp-loss}
\begin{aligned}
\mathcal{L}_{it}^{MTP} = \sum_{k=1}^{\abs{\mathcal{K}}} \mathds{1}_{k = \hat{k}} [-\textrm{log } p_{ik} + \alpha L(s^i_{t: t+N}, \hat{s}^i_{t:t+N})],
\end{aligned}
\end{equation}

\noindent where $\mathds{1}(\cdot)$ is the indicator function that equals $1$ only for the ``best matching'' mode, $k$ represents a mode, $L$ is the regression loss, and $\alpha$ is a hyper-parameter used to trade off between classification and regression. With some abuse of notation we use $\mathcal{K}$ to represent the set of trajectories predicted by a model. The original implementation~\cite{cui2019multimodal} uses a heuristic based on the relative angle between each mode and the ground truth. We select a mode uniformly at random when there are no modes with an angle below the threshold.


\noindent\textbf{Multimodal regression to anchor residuals}.
Our implementation follows the details of MultiPath (MP)~\cite{chai2019multipath}. This model implements ordinal regression by first choosing among a fixed set of anchors (computed a priori) and then regressing to residuals from the chosen anchor. The proposed per-agent loss is \eqref{eq:mtp-loss} where $\alpha = 1$ and the $k$-th trajectory is the sum of the corresponding anchor and predicted residual. To remain true to the implementation in~\cite{chai2019multipath}, we choose our best matching anchor by minimizing the average displacement to the ground truth.

We compute the set of fixed anchors by employing the same mechanism described in Section~\ref{subsec:trajectory-set-fixed}. Note that this set of trajectories is the same for all agents in our dataset. We then regress to the residuals from the chosen anchor.

\subsection{Our models}
\noindent\textbf{CoverNet (fixed)}. Our classification approach where the $\mathcal{K}$ set includes only fixed trajectories.

\noindent\textbf{CoverNet (dynamic)}. Our classification approach where the $\mathcal{K}$ set is a function of the current agent state.

\noindent\textbf{CoverNet (hybrid)}. Our classification approach where the $\mathcal{K}$ set is a combination of fixed and dynamic trajectories.

\noindent\textbf{MultiPath with dynamic anchors}. 
The MultiPath approach, extended to use dynamic anchors, described in Section~\ref{subsec:trajectory-set-dynamic}. The set of anchors is a function of the agent's speed, helping ensure that anchors are dynamically feasible. We then regress to the residuals from the chosen anchor.

\subsection{Implementation details}
Our implementation setup follows~\cite{cui2019multimodal} and~\cite{chai2019multipath}, with key differences highlighted below. See Figure~\ref{fig:network-architecture} for an overview.

We implemented our models using ResNet-50~\cite{he2015resnet} as our backbone, with pre-trained ImageNet~\cite{russakovsky2015image-net} weights downloaded from~\cite{pytorch-models}. We read the ResNet \emph{conv5} feature map and apply a global pooling layer. We then concatenate the result with an agent state vector (including speed, acceleration, yaw rate), as detailed in~\cite{cui2019multimodal}. We then add a fully connected layer, with dimension $4096$.

The output dimension of CoverNet is equal to the number of modes, namely $\abs{\mathcal{K}}$. 
For the hybrid models, the fixed:dynamic trajectory split for the nuScenes dataset is 92:682 and that of the internal dataset is 524:500. We chose these values to maximize coverage at $\epsilon \approx 2$ $meters$ and minimize the sum of the total number of categories.

For the regression models, our outputs are of dimension $\abs{\mathcal{K}} \times (\abs{\vec{x}} \times N + 1)$, where $\abs{\mathcal{K}}$ represents the total number of predicted modes, $\abs{\vec{x}}$ represents the number of features we are predicting per point, $N$ represents the number of points in our predictions, and the extra output per mode is the probability associated with each mode. For our implementations, $N = H \times F$, where $H$ represents the length of the prediction horizon in seconds, and $F$ represents the sampling frequency. For each point, we predict $(x, y)$ coordinates, so $|\vec{x}| = 2$.

Our internal datasets have $F = 10 \textrm{ } Hz$, while the publicly available nuScenes is sampled at $F = 2 \textrm{ } Hz$. We include results on two different prediction horizon lengths, namely $H = 3$ seconds and $H = 6$ seconds. 

The loss functions we use are the same across all of our implementations: for any classification losses, we utilize cross-entropy with positive samples determined by the element in the trajectory set closest to the actual ground truth in minimum average of point-wise Euclidean distances, and for any regression losses, we utilize smooth $\ell^1$. For our MTP implementation, we place equal weighting between the classification and regression components of the loss, setting $\alpha = 1$, similar to~\cite{cui2019multimodal}. 

For our classification models, we utilize a fixed learning rate of $1\mathrm{e}{-4}$. For our regression models, we use a learning rate of $1\mathrm{e}{-4}$, with a drop by $0.1$ as follows: for our internal dataset, we always perform the drop at epoch $6$; for nuScenes, we perform the drop at (1) epoch $31$ for MTP with $1$ and $3$ modes and MP dynamic with $16$ modes, (2) epoch $12$ for MTP with $16$ and $64$ modes, MP with $16$ modes and  MP dynamic with $64$ modes, and (3) epoch $7$ for MP with $64$ modes.

\subsection{Metrics}
\label{subsec:metrics}
There are multiple ways of evaluating multimodal trajectory prediction. Common measures include log-likelihood~\cite{chai2019multipath, rhinehart2019precog}, average displacement error, and hit rate~\cite{hong2019zoox}. We focus on the a) displacement error, and b) hit rate, both computed over a subset of the most likely modes.

For insight into trajectory prediction performance in scenarios where there are multiple plausible actions, we use the minimum average displacement error (ADE). The
\textrm{\ade{k}} is \( \min_{\hat{s} \in \mathcal{P}} \frac{1}{N}\sum_{\tau=t}^{t+N} || s_\tau - \hat{s}_\tau || \), where $\mathcal{P}$ is the set of $k$ most likely trajectories. We also analyze the final displacement error (FDE), which is \( || s_{t+N} - \hat{s}_{t+N}^{*} || \), where $s^*$ is the most likely mode.

In the context of planning for a self-driving vehicle, the above metrics may be hard to interpret. We use the notion of a \emph{hit rate} (see~\cite{hong2019zoox}) to simplify interpretation of whether or not a prediction was ``close enough.''
We define a $\textrm{Hit}_{k, d}$ for a single instance (agent at a given time) as 1 if \(\min_{\hat{s} \in \mathcal{P}} \max_{\tau=t}^{t+N} || s_\tau - \hat{s}_\tau || \leq d \), and 0 otherwise. When averaged over all instances, we refer to it as the $\textrm{HitRate}_{k, d}$.

\begin{figure}[t]
\begin{center}
\includegraphics[width=1.0\linewidth]{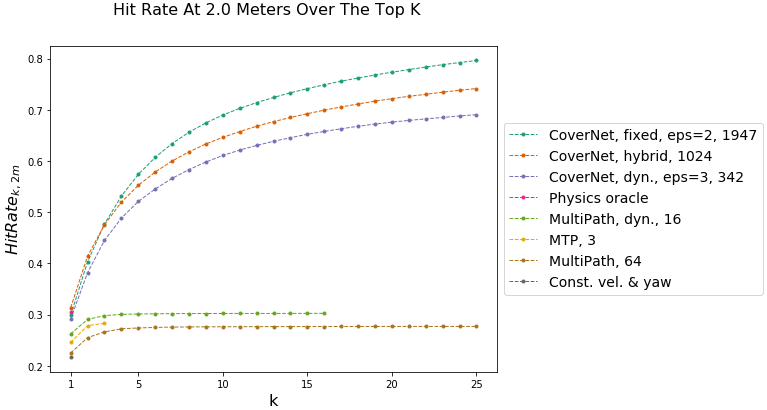}
\end{center}
\caption{Best models of each type on internal dataset (6 second horizon). CoverNet models significantly outperform others. Legend lists the model name, whether the model is dynamic or fixed (if applicable), and the number of modes.}
\label{fig:internal-hit-rate}
\end{figure}

\subsection{Input representation}
\label{sec:input}
Similar to~\cite{cui2019multimodal, chai2019multipath, djuric2018shortterm}, we rely on results from an object detection module, and we rasterize the scene for each agent as an RGB image. We start with a blank image of size ($H$, $W$, $3$) and draw the drivable area, crosswalks, and walk ways using a distinct color for each semantic category.

We rotate the image so that the agent's heading faces up, and place the agent on pixel ($l$, $w$), measured from the top-left of the image. We assign a different color to vehicles and pedestrians and choose a different color for the agent so that it is distinguishable. In our experiments, we use a resolution of 0.1 meters per pixel and choose $l = 400$ and $w = 250$. Thus, the model can ``see'' $40$ meters ahead, $10$ meters behind, and $25$ meters on each side of the agent.

We represent the sequence of past observations for each agent as faded bounding boxes of the same color as the agent's current bounding box. We fade colors by linearly decreasing saturation (in HSV space) as a function of time.

Although, we have only used one input representation in these experiments, our novel output representation can work with the input representations of~\cite{bansal2019chauffeurnet, zeng2019costmap}.

\subsection{Datasets}
\noindent\textbf{Internal self-driving dataset}.
We collected 60 hours of real-world, urban driving data in Singapore. Raw sensor data is collected by a car outfitted with cameras, lidars, and radars. A highly-optimized object detection and tracking system filters the raw sensor data to produce tracks at a 10 Hz rate. Each track includes information regarding its type (e.g., car, pedestrian, bicycle, unknown), pose, physical extent, and speed, with quality sufficient for fully-autonomous driving. We also have access to high-definition maps with semantic labels of the road such as the drivable area, lane geometry, and crosswalks.

Each ego vehicle location at a given timestamp is considered a data point. We do not predict on any tracks that are stationary over the entire prediction horizon. Our internal dataset contains around 11 million usable data points but for this analysis we created train, validation, and test sets with 1 million, 300,000, and 300,000 data points, respectively.

\bigskip

\noindent\textbf{nuScenes}.
\label{subsec:nuscenes}
We also report results on nuScenes~\cite{nuscenes2019}, a public self-driving car dataset. nuScenes consists of 1000 scenes, each 20 seconds in length. Scenes are taken from urban driving in Boston, USA and Singapore. Each scene includes hand-annotated tracks and high-definition maps. Tracks have 3D ground truth annotations, and are published at 2 Hz. Since annotations are not public on the test set, we created a set for validation from the train set (called the train-val set) and treated the validation set as the test set. As with our internal dataset, we removed vehicles that are stationary and also removed vehicles that go off the annotated map. This leaves us with 32,186 observations in the train set, 8,560 observations in the train-val set, and 9,041 observations in the validation set. This split publicly available in the nuScenes \href{https://www.nuscenes.org/}{software development kit}~\cite{nuScenes-website}.

\begin{table}[t]
\centering
\begin{tabular}{p{1.2cm}p{1.2cm}p{1.2cm}p{1.2cm}p{1.2cm}}
\toprule
Method  & \ade{1} & \ade{5} & \ade{10} & \ade{15} \\
\midrule
$\text{max } \ell^2$ & 1.0    &  0.67   &    0.64    & 0.64         \\
$\text{average } \ell^2$ &  0.96   &  0.66   &  0.64      &   0.64       \\
$\text{RMS of }  \ell^2$ &  0.96   &  0.66  &  0.64       &  0.63        \\
\bottomrule
\end{tabular}
\caption{Ground truth matching for fixed trajectory set (150 modes) on internal dataset (3 sec horizon).}
\label{tab:ablation-ground-truth}
\end{table}






\section{Results}
The main results are summarized in Table~\ref{tab:main-results}. Qualitative results are shown in Figure~\ref{fig:image-examples}.

\begin{table*}
\small
\centering
\begin{tabular}{llcccccc}
\toprule
Method & Modes   & \ade{1} $\downarrow$ & \ade{5} $\downarrow$ & \ade{10} $\downarrow$ & \ade{15} $\downarrow$ & FDE $\downarrow$ & \hitrate $\uparrow$ \\
\midrule 
Const. vel. \& yaw & N/A   & 4.61 (3.63) & 4.61 (3.63) & 4.61 (3.63) & 4.61 (3.63) & 11.21 (9.86) & 0.09 (0.22)  \\
Physics oracle &  N/A             & \textbf{3.70} (\textbf{1.88}) & 3.70 (1.88) & 3.70(1.88) & 3.70 (1.88) & \textbf{9.09} (5.72) & 0.12 (0.31)  \\
\midrule 
MTP~\cite{cui2019multimodal}  &   1 (1)          & 4.17 (\textbf{1.88}) & 4.17 (1.88) & 4.17 (1.88) & 4.17 (1.88) & 9.37 (\textbf{5.22}) & 0.05 (0.24)  \\
MTP~\cite{cui2019multimodal}  &   3 (3)          & 4.13 (2.01) & 2.93 (1.73) & 2.93 (1.73) & 2.93 (1.73) & 9.23 (5.45) & 0.10 (0.28)  \\
MTP~\cite{cui2019multimodal}  &   16 (16)         & 4.55 (3.15) & 3.32 (2.48) & 3.25 (2.43) & 3.23 (2.42) & 9.58 (7.79) & 0.08 (0.25)  \\
MTP~\cite{cui2019multimodal}  &   64 (64) & 4.50 (3.21) & 3.24 (2.63) & 3.15 (2.51) & 3.13 (2.47) & 9.59 (7.74) & 0.09 (0.27)  \\
MultiPath~\cite{chai2019multipath} &   16 (16)   & 4.89 (2.34) & 2.64 (1.71) & 2.47 (1.71) & 2.43 (1.70) & 10.41 (5.83) & 0.08 (0.24)  \\
MultiPath~\cite{chai2019multipath} &   64 (64)   & 5.05 (2.30) & 2.32 (1.42) & 1.96 (1.36) & 1.86 (1.34) & 10.69 (5.63) & 0.10 (0.27)  \\
MultiPath~\cite{chai2019multipath}, dyn. & 16 (16) & 3.89 (2.06) & 3.34 (1.47) & 3.28 (1.46) & 3.27 (1.46) & 9.19 (5.76) & 0.10 (0.30)  \\
MultiPath~\cite{chai2019multipath}, dyn. & 64 (64) & 4.05 (2.23) & 3.45 (1.53) & 3.33 (1.46) & 3.28 (1.44) & 9.47 (6.17) & 0.13 (0.28)  \\
\midrule 
CoverNet, fixed, $\varepsilon$=8 & 64 (64) & 5.16 (2.77) & 2.41 (1.98) & 2.18 (1.93) & 2.13 (1.93) & 10.84 (6.65) & 0.08 (0.06)  \\
CoverNet, fixed, $\varepsilon$=5 & 232 (208) & 4.73 (2.32) & 2.14 (1.35) & 1.72 (1.25) & 1.60 (1.22) & 10.16 (5.67) & 0.15 (0.31)  \\
CoverNet, fixed, $\varepsilon$=4 & 415 (374) & 5.07 (2.27) & 2.31 (1.29) & 1.76 (1.15) & 1.57 (1.10) & 10.62 (5.85) & 0.17 (0.35)  \\
CoverNet, fixed, $\varepsilon$=3 & 844 (747) & 4.74 (2.28) & 2.32 (1.32) & 1.74 (1.13) & 1.51 (1.07) & 10.19 (5.92) & 0.23 (0.33)  \\
CoverNet, fixed, $\varepsilon$=2 & 2206 (1937) & 5.41 (2.16) & 2.62 (\textbf{1.16}) & 1.92 (\textbf{0.93}) & 1.63 (\textbf{0.84}) & 11.36 (5.53) & 0.24 (\textbf{0.57})  \\
CoverNet, dyn., $\varepsilon$=3 & 357 (342) & 3.90 (2.06) & 2.02 (1.17) & 1.57 (0.97) & 1.36 (0.88) & 9.65 (5.90) & 0.33 (0.52)  \\
CoverNet, hybrid & 774 (1024) & 3.87 (2.18) & \textbf{1.96} (1.24) & \textbf{1.48} (0.99) & \textbf{1.28} (0.88) & 9.26 (5.84) & \textbf{0.33} (0.55)  \\
\bottomrule
\end{tabular}
\caption{nuScenes and internal datasets (6 sec horizon). Results listed as nuScenes (internal). Smaller minADE\textsubscript{k} and FDE is better. Larger \hitrate is better. Dyn. = dynamic, vel. = velocity, const. = constant, $\varepsilon$ is given in meters.}
\label{tab:main-results}
\end{table*}

\bigskip 

\noindent\textbf{Quantitative results}. Across the six metrics and the two datasets we used, CoverNet outperforms previous methods and baselines in 8 out of 12 cases.
However, there are big differences in method ranking depending on the metric.

CoverNet represents a significant improvement on the \hitrate metric, achieving $33\%$ on nuScenes with the hybrid trajectory set. The next best model is MultiPath, where our dynamic grid extension is a slight improvement over the fixed grid used by the authors ($13\%$ vs. $10\%$). MTP with three modes performs worse, achieving $10\%$, barely outperforming the constant velocity baseline.

We notice a similar pattern on the internal dataset, where CoverNet outperforms previous methods and baselines. 
Here, the fixed set with 1,937 modes performs best ($57\%$), closely followed by the hybrid set ($55\%$).
Among previous methods, again MultiPath with dynamic set works the best at $30\%$ \hitrate. Figure~\ref{fig:internal-hit-rate} shows that CoverNet significantly outperforms previous methods as the hit rate is expanded over more modes.

CoverNet also performs well according the Average Displace Error \ade{k} metrics, in particular for $k \in \{5, 10, 15\}$, where we see CoverNet outperforming state-of-the-art methods in every category.
Most notably, under the \ade{15} metric for our internal dataset, the hybrid CoverNet with fixed set and 2,206 modes performs best with \ade{15} of $0.84$, 4x better than the constant velocity baseline and 2x better than the MTP and MultiPath.
For the \ade{1} metric the regression methods performed the best.
This is not surprising since for low $k$ it is more important to have one trajectory very close to the ground truth, a metric paradigm that favors regression over classification.

A notable difference between nuScenes and internal is that the \hitrate and \ade{k} continues to improve for larger sets, while it plateaus, or even decreases at around 500-1,000 modes on nuScenes.
We hypothesize that this is due to relatively limited size of nuScenes.

\bigskip

\noindent\textbf{Qualititive results}.
In Figure~\ref{fig:image-examples}, we show the visualization of a scene overlaid with predictions from our top models compared against our baselines. We note that our prediction horizon for this scene is six seconds. As such, the predictions do not reflect collisions as the pedestrians in the scene will have crossed the road before our vehicle reaches the pedestrian pose reflected in the images. 

We emphasize that the CoverNet predictions do not include straight trajectories
because the vehicle slows down before the curve. When visualized as a video, we first predict straight trajectories, followed by predicting left turn trajectories when the vehicle starts slowing down. We highlight the smoothness of the trajectories predicted by our model contrasted against the regression baselines. Figure~\ref{fig:image-examples} also suggests that the different alternatives for left turns are better captured by CoverNet than by the baseline models.


\section{Ablation studies}

\subsection{Distance function}
\label{subsec:ablation-ground-truth}

We analyzed different methods for matching the ground truth to the most suitable trajectory in the trajectory set.
Table~\ref{tab:ablation-ground-truth} compares performance using the 
max, average, and root-mean-square of the point-wise error vector of Euclidean distances
for matching ground truth to the ``best'' trajectory in a fixed trajectory set of size 150. 
Performance is relatively consistent across all three choices, so we picked the average point-wise $\ell^2$ norm to better align with related regression approaches~\cite{chai2019multipath}.

\subsection{Dynamic vs fixed trajectory set coverage}

In Figure~\ref{fig:epsilon-coverage}, we compare the number of trajectories needed to achieve 100\% coverage of the trajectory set for different levels of $\varepsilon$ for the fixed and hybrid trajectory set generation functions, where the latter use a mix of fixed and dynamic trajectories.
This figure highlights the advantage of adding dynamic trajectories: they are able to achieve the same level of coverage as the fixed trajectories, but need a smaller number of trajectories to do so.


\vspace{-2mm}
\section{Conclusion}
\vspace{-2mm}
We introduced CoverNet, a novel method for multimodal, probabilistic trajectory prediction in real-world, urban driving scenarios.
By framing this problem as classification over a diverse set of trajectories, we were able to a) ensure a desired level of coverage of the state space, b) eliminate dynamically infeasible trajectories, and c) avoid the issue of mode collapse.
We showed that the size of our trajectory sets remain manageable over realistic prediction horizons.
Dynamically generating trajectory sets based on the agent's current state further improved performance.
We compared our results to multiple state-of-the-art methods on real-world self-driving datasets (public and internal), and showed that it outperforms similar methods.

\noindent\textbf{Acknowledgments}. We would like to thank Emilio Frazzoli and Sourabh Vora for insightful discussions, and Robert Beaudoin for help on the implementation.

{\small
\bibliographystyle{ieee_fullname}
\bibliography{../bibliography}
}

\clearpage

\appendix
\title{Supplementary Material}

\section{Trajectory set visualization}

\begin{figure*}
\begin{center}
\includegraphics[width=0.65\linewidth]{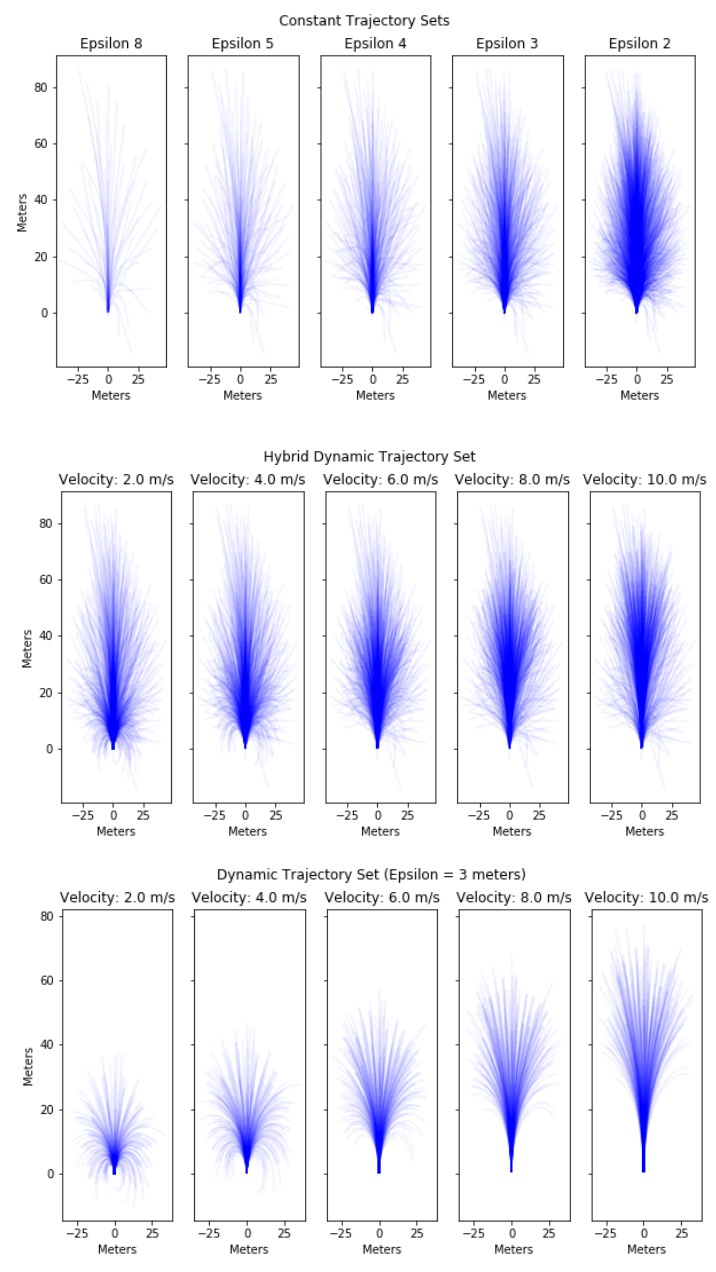}
\end{center}
\vspace{-3mm}
\caption{Visualization of fixed and dynamic trajectory sets for different coverage levels on the internal dataset for a $6$ second prediction horizon.}
\label{fig:trajectory-set-visualization}
\end{figure*}

Figure~\ref{fig:trajectory-set-visualization} visualizes the trajectory sets for fixed, hybrid, and dynamic trajectory set generation functions to better understand the space CoverNet learns to classify over.
We present the fixed trajectory sets for different coverage levels and the dynamic trajectory sets for a fixed coverage level but varying speeds.

\section{\ade{1} over the prediction horizon}

\begin{figure}[H]
\begin{center}
\includegraphics[width=1.0\linewidth]{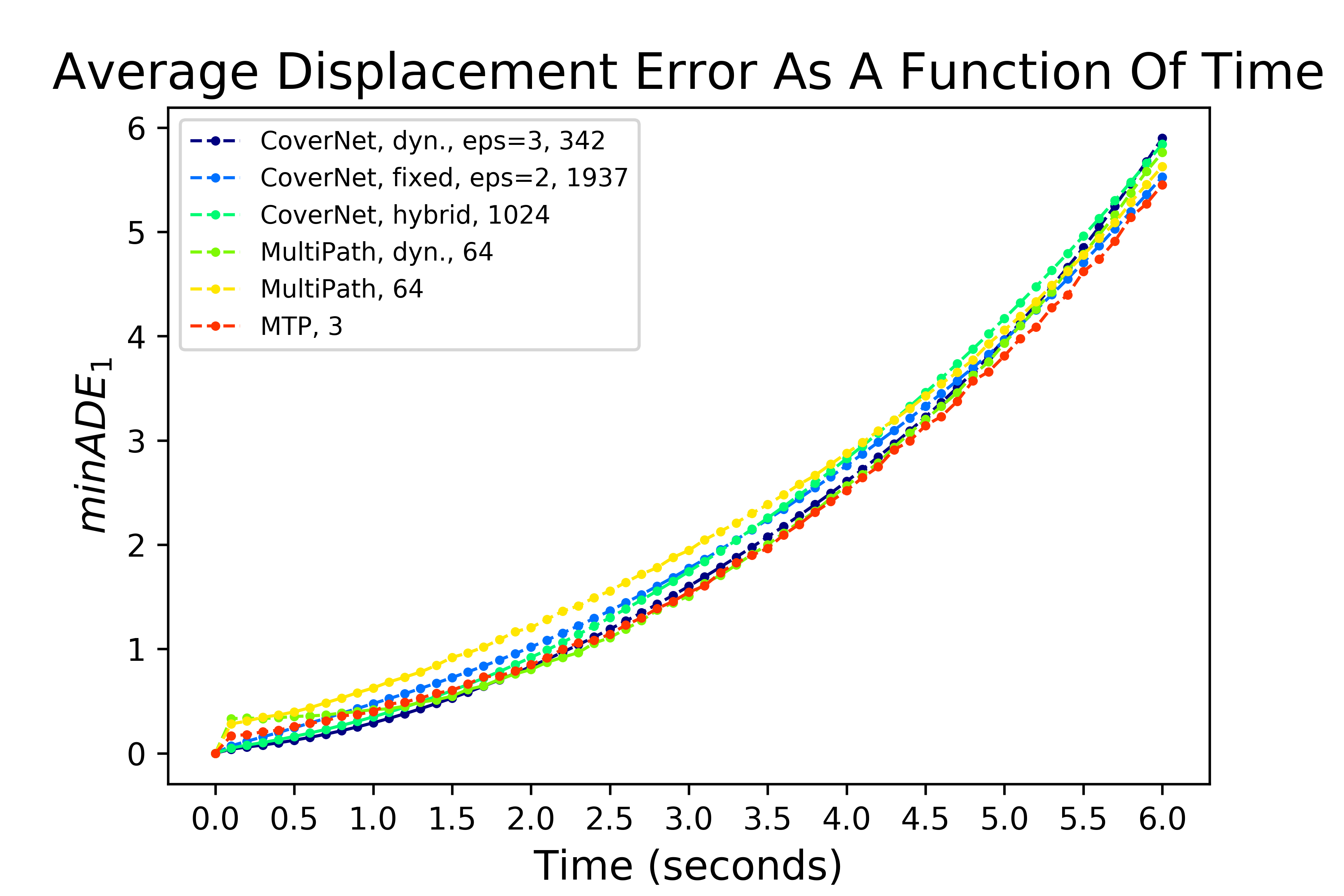}
\end{center}
\vspace{-3mm}
\caption{The \ade{1} over the prediction horizon for the best regression and CoverNet models on our internal dataset.}
\label{fig:min-ade-over-horizon}
\end{figure}

In Figure ~\ref{fig:min-ade-over-horizon}, we show the \ade{1} for the best regression and CoverNet models to demonstrate how the performance of the different methods scales with time. We note that all of the methods show similar performance over time for the most likely mode.

\section{Greedy Approximation Algorithm}

In~\ref{subsec:trajectory-set-fixed}, we detail our \emph{bagging} algorithm. This is a greedy approximation algorithm for solving $\eqref{eq: set-cover}$. 

In this work, we utilize a deterministic variant of this algorithm, detailed in~\ref{subsec:trajectory-set-fixed}. We mention that another possibility is to use a random version of this algorithm. This variant makes a weighted random choice based on how many uncovered trajectories the candidates are covering until there are no more trajectories to cover. This can be repeated many times to obtain multiple bags, and we can choose the cover set based on the smallest number of elements.

\end{document}